\documentclass{egpubl}
\usepackage{IMET2023}

 \usepackage{booktabs}
 \usepackage{multirow}
 
\WsPaper           % uncomment for final version of EG Workshop contribution
\usepackage[T1]{fontenc}
\usepackage{dfadobe}  

\usepackage{cite}  % comment out for biblatex with backend=biber
\BibtexOrBiblatex
\electronicVersion
\PrintedOrElectronic
\ifpdf \usepackage[pdftex]{graphicx} \pdfcompresslevel=9
\else \usepackage[dvips]{graphicx} \fi
\usepackage{orcidlink}
\usepackage{egweblnk} 
\title{Fictional Worlds, Real Connections: Developing Community Storytelling Social Chatbots through LLMs}

\author[Sun et al.]
{\parbox{\textwidth}{\centering {\footnotesize Yuqian Sun}$^{1}${\Large\orcid{0000-0002-4076-8140}}, 
{\footnotesize Hanyi Wang}$^{2}${\Large\orcid{0009-0009-1871-0725}}, 
{\footnotesize Pok Man Chan}$^{3}$, 
{\footnotesize Morteza Tabibi}$^{4}${\Large\orcid{0009-0005-1309-4573}}, 
{\footnotesize Yan Zhang}$^{4}${\Large\orcid{0009-0002-9466-6534}}, 
{\footnotesize Huan Lu}$^{5}${\Large\orcid{0009-0003-2717-1926}}, 
{\footnotesize Yuheng Chen}$^{4}${\Large\orcid{0009-0001-9013-7312}}, 
{\footnotesize Chang Hee Lee}$^{6}${\Large\orcid{0000-0002-5384-5547}}, 
{\footnotesize Ali Asadipour}$^{7}${\Large\orcid{0000-0003-0159-3090}},
	}
	\\
 	\\
	{\parbox{\textwidth}{\centering 
 $^1$ Computer Science Research Centre, Royal College of Art, United Kingdom\\
 $^2$ Independent, United States\\
 $^3$ Independent, Hong Kong\\
 $^4$ rct.ai, United States\\
 $^5$ Independent, China\\
 $^6$ Affective Systems and Cognition Lab, Industrial Design Department, College of Engineering, KAIST, United Kingdom\\
 $^7$ Royal College of Art, United Kingdom\\
		}
	}
}

\begin{document}
	
	% uncomment for using teaser
	% \teaser{
	%  \includegraphics[width=\linewidth]{eg_new}
	%  \centering
	%   \caption{New EG Logo}
	% \label{fig:teaser}
	%}
	
	\maketitle
	%-------------------------------------------------------------------------
	\begin{abstract}
	We address the integration of storytelling and Large Language Models (LLMs) to develop engaging and believable Social Chatbots (SCs) in community settings. Motivated by the potential of fictional characters to enhance social interactions, we introduce \textit{Storytelling Social Chatbots} (SSCs) and the concept of \textit{story engineering} to transform fictional game characters into "live" social entities within player communities.
	Our story engineering process includes three steps: (1) Character and story creation, defining the SC's personality and worldview, (2) Presenting Live Stories to the Community, allowing the SC to recount challenges and seek suggestions, and (3) Communication with community members, enabling interaction between the SC and users. We employed the LLM GPT-3 to drive our SSC prototypes, ``David" and ``Catherine," and evaluated their performance in an online gaming community, ``DE (Alias)," on Discord.
	Our mixed-method analysis, based on questionnaires (N=15) and interviews (N=8) with community members, reveals that storytelling significantly enhances the engagement and believability of SCs in community settings.
	
	\textbf{Keywords}: social chatbot, believable agent
	
%		\begin{CCSXML}
%			<ccs2012>
%			<concept>
%			<concept_id>10010147.10010371.10010352.10010381</concept_id>
%			<concept_desc>Computing methodologies~Collision detection</concept_desc>
%			<concept_significance>300</concept_significance>
%			</concept>
%			<concept>
%			<concept_id>10010583.10010588.10010559</concept_id>
%			<concept_desc>Hardware~Sensors and actuators</concept_desc>
%			<concept_significance>300</concept_significance>
%			</concept>
%			<concept>
%			<concept_id>10010583.10010584.10010587</concept_id>
%			<concept_desc>Hardware~PCB design and layout</concept_desc>
%			<concept_significance>100</concept_significance>
%			</concept>
%			</ccs2012>
%		\end{CCSXML}
%		
%		\ccsdesc[300]{Computing methodologies~Collision detection}
%		\ccsdesc[300]{Hardware~Sensors and actuators}
%		\ccsdesc[100]{Hardware~PCB design and layout}
%		
%		
%		\printccsdesc   
	\end{abstract}  
	%-------------------------------------------------------------------------
\section{Introduction}

\begin{figure*}[htbp]
	\centering 
	\vspace{-1em}
	\includegraphics[width=1.03\textwidth]{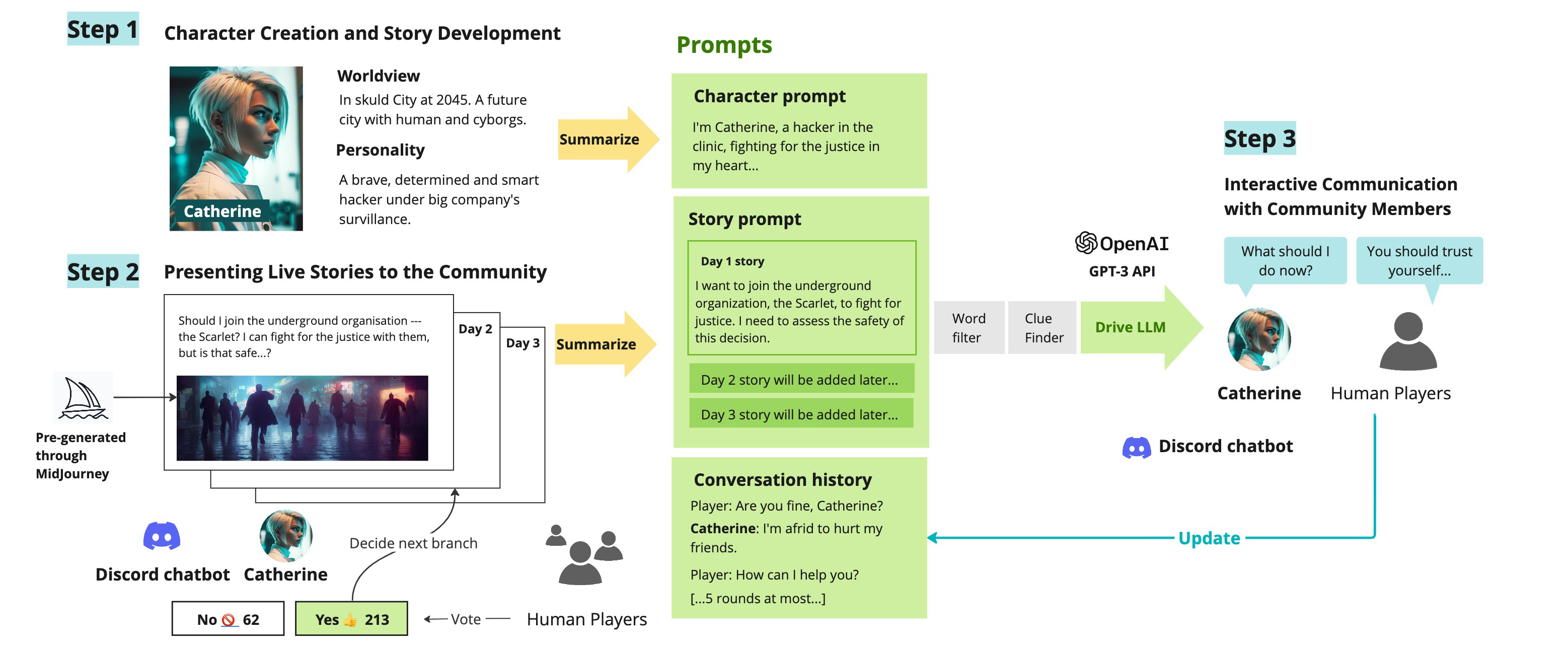} 
	\caption{Story engineering for Storytelling Social Chatbot (SSC) Catherine} 
	\label{fig:storyEngineering} 
	\vspace{-1.5em}
\end{figure*}

Fictional characters are central to the research of social chatbots: artificial intelligence (AI) dialogue systems capable of having social and empathetic conversations with users \cite{brandtzaegMyAIFriend2022}. The field of Social Chatbots (SCs) has been rapidly advancing with the development of Large Language Models (LLMs), which enable SCs to generate dynamic and personalized responses for different users. These AI dialogue systems are becoming increasingly human-like and are taking on a variety of social roles such as assistant, partner, friend\cite{metz_2020}\cite{pentina_hancock_xie_2023}, citizens\cite{park2023generative}, and YouTuber. While researchers have begun studying SCs in relation to Human-AI friendship and intimate relationship development, most studies have focused on interactions with a single bot rather than in richer social contexts. Furthermore, existing research has not deeply considered the role of fictional content. To some extent, SCs are similar to game characters as they are both fictional entities. Computing systems are wrapped into different``people" with distinct names, personalities, and background stories. As such, the question remains: Can fictional characters live together with humans, and become members of human society? How will their fictional identities and stories influence our perception?

Previous researchers explored chatbots as community members using a Twitch chatbot \cite{seering_luria_ye_kaufman_hammer_2020}, but their study did not employ LLMs, which offer more advanced language abilities, necessitating further investigation. Additionally, current LLM-driven SC like replika\cite{brandtzaegMyAIFriend2022} does not focus on multi-person social contexts. By integrating insights from gaming research, particularly non-player characters (NPCs)\cite{bowman_2019}, we strive to develop a better understanding of how SCs can become more believable and engaging within community settings.

We explore the idea that fictional characters can share their lives as social chatbots in the human community context, just like friends sharing their lives on social media and receiving feedback. Rich social interactions, including conversations with the character, seeing the character's interactions with other people, and the dynamics in the character's life, can make the character believable and the interaction engaging and meaningful.

Motivated by these views, we investigate the following research questions:

\textbf{RQ1} - How can an LLM-based SC enhance engagement and meaningful interactions within a community setting?

\textbf{RQ2} - How does the integration of fictional stories in the SC's design impact its believability and engagement?

We refer to the process of transforming a fictional character into a social chatbot as \textit{story engineering}. This term is adapted from \textit{prompt engineering}, which focuses on designing text prompts to influence the generated content of an LLM. In contrast, designing an LLM-based ``live" character with social behaviour requires considering multiple aspects: the character's personality and story, the LLM's generation goals, and the interaction methods facing people. Our prototype, Storytelling Social Chatbot (SSC), implements this concept through the following processes:

(1) Story and character design, where we define the SC's personality and the worldview they inhabit, (2) Presenting Live Stories to the Community, allowing the SC to recount challenges and problems they need suggestions on, and (3) Communication with community members, enabling community members to chat with the SC. We designed a workflow to drive the character through the LLM GPT-3.

To evaluate our workflow, we introduced two fictional characters, ``David" and ``Catherine," as SSCs in an online gaming community ``DE (Alias)" on Discord. These characters followed the worldview and story of the under-development game``DE". Every day, the character shared their current situation in the story channel (e.g., ``I'm chased by the evil agent, what should I do?"), similar to social media. Meanwhile, community members could engage with them and discuss their current situation in the chat channel, suggesting decisions through voting on the character's choice. At the end of the day, the character made their decision based on voting and released the next story.

The DE gaming community already had a non-storytelling chatbot based on LLM GPT-3\cite{OpenAIAPIa}, Jerry, which we used as a benchmark to evaluate the effectiveness of SSCs. We collected qualitative feedback through questionnaires with 15 core community members and interviews with eight of them. Our mixed-method analysis reveals that storytelling enhances the engagement and believability of SCs in community settings. We summarize the themes related to SSC and discuss the design implications for future developments to make SC interactions more engaging and meaningful.

In summary, we contribute (1) Two community-based SSCs developed through our concept of story engineering based on GPT-3, and (2) Insights from their development and evaluation with community members. By designing SSCs with attention to the specific social context and using storytelling to enhance their believability and engagement, we can create a new generation of SSCs that can contribute to our social lives in novel and exciting ways.

%\begin{figure}[htbp]
%\centering 
%\includegraphics[width=0.5\textwidth]{figure/cathe call.jpg} 
%\caption{Concept art for SC(Catherine) calling players for help.} 
%\label{fig:cathCall} 
%\end{figure}

\section{Background}

In online communities, achieving meaningful interaction is essential. Although many active communities employ bots with both moderating and entertaining functions\cite{discord}, their interactions may not necessarily generate new information or contribute to the community's values or project goals. Narratives, according to Ricoeur, are vital for constructing our sense of self, making sense of our experiences, and creating meaning in our lives\cite{albrechtslund_2010}. Storytelling has been considered a means and approach in various contexts, such as children's education\cite{zhang_xu_wang_yao_ritchie_wu_yu_wang_li_2022}, \cite{jackson_latham_2022}, healthcare, and skill learning\cite{suh_zhao_law_2022}. In online community building, particularly game-related communities, storytelling may serve as a crucial goal or primary means of maintaining community activity and a sense of meaning\cite{albrechtslund_2010}.

Integrating AI as a member of an online community could potentially become a source of discussion or meaning by sharing its life experiences consistent with the content the entire community focuses on. These experiences are unlike the small talk generated by existing SCs such as Replika\cite{brandtzaegMyAIFriend2022}. While there has been extensive research on dyadic interaction chatbots%\cite{brandtzaeg_skjuve_følstad_2022}
\cite{brandtzaeg_skjuve_folstad_2022}
\cite{chaves_gerosa_2020}\cite{shin_yoon_kim_lee_2021}, multi-party chatbots, particularly those acting as community members, remain underexplored.

Researchers have developed a Twitch chatbot that focuses on the social context of a community, making the chatbot a social member\cite{seering_luria_kaufman_hammer_2019}. However, this does not consider the potential impact of current LLMs, such as uncontrollable content generation and human attachment. Seering et al. proposed various ideas for community chatbots, including the storyteller bot concept\cite{seering_luria_kaufman_hammer_2019}. They suggested that a more interactive, almost ``live" narrative experience could be created by having chatbots that are regular community members involved with other chatbots in engaging ways. Furthermore, chatbots should focus on being deployed in specific social contexts.

Developing social abilities in chatbots requires mimicking human behaviour to some extent. Accordingly, previous research on SCs has focused on personification, emotions\cite{heudin_2018}, and other related aspects. Similarly, research on NPCs' social behaviour in games addresses emotional attachment\cite{emmerich_ring_masuch_2018}\cite{bopp_müller_aeschbach_opwis_mekler_2019}\cite{bowman_2019}, empathy\cite{chaves_gerosa_2020}, and identity\cite{dimas_1970}. Due to the multimodal nature of video game experiences, the influencing factors in related research are more diverse, such as character appearance and game environment. Overall, computer games can be considered a social ``training ground" for NPCs\cite{bowman_2019}, and natural language interaction opens room for players to provide new content that can be (to an extent) acknowledged by the game\cite{park2023generative}.

Since we focus on character dialogues in a story context, we focused on believability. Anton et al. defined believability as the extent to which users interacting with the agent come to believe that they are observing a sentient being with its own beliefs, desires, and personality\cite{bogdanovychWhatMakesVirtual2016a}. The authors concluded that a believable character is not necessarily a real character but must be real in the context of its environment. Virtual agents that can adapt to changes in the environment and exist in the correct social context are those perceived as more believable. Kiran et al. argues that the believability of such agents is tightly connected with their ability to relate to the environment during a conversation\cite{ijazEnhancingBelievabilityEmbodied}. Believable characters can create better player experiences, and accordingly, believable SC should bring a more engaging experience to a community.

Considering that online community interactions are mainly text-based and do not involve 3D scenes or embodied agents, we synthesized several scholars' discussions and regarded \textbf{emotions, personality, and motivation} as the criteria for measuring SSCs believability. By focusing on these aspects, we can aim to design a storytelling SC that is logical, coherent, and clear, leading to a new generation of SCs that contribute to our social lives in innovative ways.

\begin{table*}
	\centering
	\caption{Character and Stage Overview}
	\label{characterList}
	\begin{tabular}{|p{1.5cm}|p{4.5cm}|p{3.2cm}|p{7.5cm}|}
		\hline
		\textbf{Character and Stage} & \textbf{Function} & \textbf{Duration} & \textbf{Background story} \\
		\hline
		\multirow{2}{*}{\parbox{2cm}{Jerry: \\ Benchmark}} & \multirow{2}{3.5cm}{Chat with people.} & Available 4 months later after the community opened in 2021 December. & A traveller in the metaverse. \\
		\hline
		\multirow{4}{*}{\parbox{2cm}{David: \\ Pilot Study}} & \multirow{4}{3.5cm}{Releasing stories and chatting with members through the adventure.} & Opened through the 3 days warm-up, 3 days main stories in 2022 November. & An employee in a pharmacy institute ``BioTech" under the big company ``Domain". Ran away after finding the illegal trade in the company. Met and saved by Catherine and her father, who turned out to be the hidden villain that made Catherine mind-controlled.  \\
		\hline
		\multirow{4}{*}{\parbox{2cm}{Catherine: \\ Main Study}} & \multirow{4}{3.5cm}{Same as David plus clue finder function: replying to specific questions with pre-scripted replies.} & Opened through the 1-day warm-up, 3 days main stories at 2023 February. & Under surveillance by the big company ``Domain" who used to mind-control her but released her after David's story. A hacker and doctor in the clinic. Wants revenge on ``Domain". Regards David as a big brother. \\
		\hline
	\end{tabular}
	\vspace{-1.5em}
\end{table*}

\section{SSC Development and Implementation in the Community Context}

In this section, we will discuss the development of the Storytelling Social Chatbots (SSCs), ``David" and ``Catherine," within the context of the DE gaming community. We will cover the background of the DE game, the DE community, the character background and story settings, and the detailed experimental workflow.

\subsection{DE Game and Community Background}

The DE game is an under-development first-person shooting player vs player (PVP) game set in a futuristic virtual world featuring digitized real humans. Players engage in battles for interests, power, and resources, shaping the future of the virtual world. The game is based on blockchain technology, and its community members are open to new technologies. The DE community has a large number of members on Discord (n=97841), with 7006 having sent over 5 messages in the community. However, it should be noted that in Web3 and blockchain-based Discord gaming communities, a considerable amount of members are inactive after joining. Typically, players adopt a wait-and-see approach after encountering project promotions before deciding whether to follow up. The active membership ranges from around 100-200 people, with the most core members being around 15-20 who can directly discuss moderating issues with official game team members. Furthermore, the DE community already had a non-storytelling LLM-based chatbot, Jerry, which we used as a benchmark to evaluate the effectiveness of SSCs. We provide a brief overview of their story backgrounds and purposes in \ref{characterList}.

\subsection{Character Background and Story Settings}

After consulting with the DE game's development team, we selected two future in-game characters, David and Catherine, for our research experiment. David is a doctor who uncovers illegal operations involving mind control technology within the company he works for and decides to escape with crucial information. Catherine is a skilled hacker who discovers her past is filled with tragedy, fueling her hatred for oppressive forces and motivating her to fight against them.

The development and story settings of these two characters in our research are consistent with the game's official narrative. The experiment conducted in this research also serves as a player community activity, promoting the game's characters and worldviews while engaging the community.

\subsection{Detailed Experimental Workflow}

To develop and implement the SSCs within the community context, we followed a three-step workflow as part of our story engineering concept:

\begin{enumerate}
	\item \textbf{Character Creation and Story Development}: Define the characters' personalities, backgrounds, and stories, as well as the world they inhabit.
	
	\item \textbf{Presenting Live Stories to the Community}: Share the characters' live stories with the community, allowing them to interact and provide suggestions. Each day, the character will share their current situation in the \textit{story channel}, with community members who can vote on the character's choices. Concurrently, community members can engage with the SSCs and discuss their current situations in the \textit{chat channel}. Based on the voting results, the character will make a decision and release the next part of the story at the end of the day.  
	
	\item \textbf{Interactive Communication with Community Members}: Enable community members to communicate with the SSCs, discussing their situations. Each day, the character will change their state upon the current live story, and continue to talk to community members and respond to their suggestions throughout the day.
\end{enumerate}

We conducted a pilot study with David's character in November 2022 to preliminarily assess the efficacy of SSCs. Subsequently, we treated Catherine's character as the main study, conducting questionnaires and interviews with core community members, and asking them to compare Catherine and Jerry to investigate the impact of storytelling on SCs.

\section{Implementation of Story Engineering}

In this section, we will explain the different stages of story engineering Fig.\ref{fig:storyEngineering}, focusing on the creation and implementation of SSCs ``David" and ``Catherine." 

\subsection{Step 1: Character Creation and Story Development}

To design the story of characters, the following perspectives should be considered:

\textbf{Worldview}: Determine the setting of the story. Consider the rules, customs, and culture of this world, as they will influence the character's interactions with the community. For example, following the background story of DE, stories happen in Skuld City in 2045, a futuristic city inhabited by humans and cyborgs.

\textbf{Personality}: Define the character's personality and motivations. Consider how these characteristics will shape their decisions and behaviour throughout the story. For instance, Catherine is a brave, determined, and smart hacker, and she tries to fight for justice against a large company's control.

\subsection{Step 2: Presenting Live Stories to the Community}

The story prepared by the team members will be posted by the SSC in the story channel, similar to a short novel, and should consider the reading experience of the players. 

The current situation the character is experiencing should be identified, including their goals, motivations, and the challenges they face. This will provide a context for the character's interactions with the community and the decisions they need to make. For example, David will say: ``Today, my world turned upside down when I got shot by Catherine, who I always trust..." At this point, he needs to decide whether to continue trusting his friend. The design of this part is similar to branching choices in traditional interactive narrative video games, requiring the scriptwriter to plan each branch in advance.

In our project, although the story is mainly conveyed through text, we also prepared illustrations for easier understanding by the audience. The illustrations were generated by MidJourney\cite{MidJourney} before the activity started based on the story. Additionally, the character's decision options were presented with representative emoji reactions as selectable buttons.

Using the discord.py\cite{discord} tool provided by Discord official team, we can schedule the daily live story posts, including text, images, and options. Players can vote on the options and view the current voting results. On the next day, following the higher-voted choice, the next story will be published.

To better introduce the character, we put the beginning days of the event as a warm-up phase, the character's story didn't provide a choice. After that, players will get a better understanding of the character, so they can be more considerate when they talk to the character.

\subsection{Step 3: Interactive Communication with Community Members}

To enable interactive communication with community members, we employed a conversation system consisting of several modules that work together to drive the chatbot's dialogue system(Fig.\ref{fig:storyEngineering}):

\subsubsection{Words Filter}

This module is designed to filter out offensive inputs/outputs by scanning the text through a list of keywords.

\subsubsection{Clue Finder (Catherine only)}
\label{sec:clue}
The Clue Finder is a small API microservice designed for providing static replies. It compares a given sentence to a list of keywords using the SentenceTransformer\cite{reimers-2019-sentence-bert} model. When a match is found, it returns the corresponding image URL and text. The service is lightweight and easy to deploy in a private network, making it suitable for projects that require fast responses with low latency.

When the input is similar to options in keywords (e.g., ``Give me some information about Domain"), Catherine will reply with the pre-written text and won't call the OpenAI service. This function saves time and expense for basic questions and explains the worldview with several options, including images.

\subsubsection{LLM Configuration}

When the Clue Finder does not generate a response, the OpenAI GPT-3's API is called to generate a reply based on the given prompt. The prompt consists of three parts: character prompt, live story prompt, and dialogue history.

\textbf{Character prompt}: Provides background information on the character (e.g., ``I'm David, the doctor in the clinic.") Adapted from the character information prepared in Step 1. To make the current SSC understand the current story and save words in the prompt, some simplifications and adaptations are needed.

\textbf{Live Story prompt}: Derived from Step 2, it conveys the current situation or context to the LLM, enabling it to generate contextually appropriate responses. As the story progresses, the story prompt will also become longer.

\textbf{Dialogue history}: This consists of chat records from the past five rounds of conversation, allowing the character to develop brief memories and respond coherently to ongoing discussions.

This conversation system allows the SSCs to communicate with community members, respond to their suggestions, and adapt their stories based on the current live story. 

\section{Pilot study: David}

For the pilot study, we first introduced a character named ``David" in 2022 November. We created a story about how he discovers the dark side of Skuld City (the main stage of DE) and how he fights back against the villainous company ``Domain". The main goal of the pilot study with David is to validate the feasibility of storytelling SCs and their attraction within the community. We released David during the DE game testing days, which attract more players in the community.

\subsection{Observations on Made-up content from LLM}

\begin{figure}[htbp]
	\centering 
	\vspace{-1em}
	\includegraphics[width=0.42\textwidth]{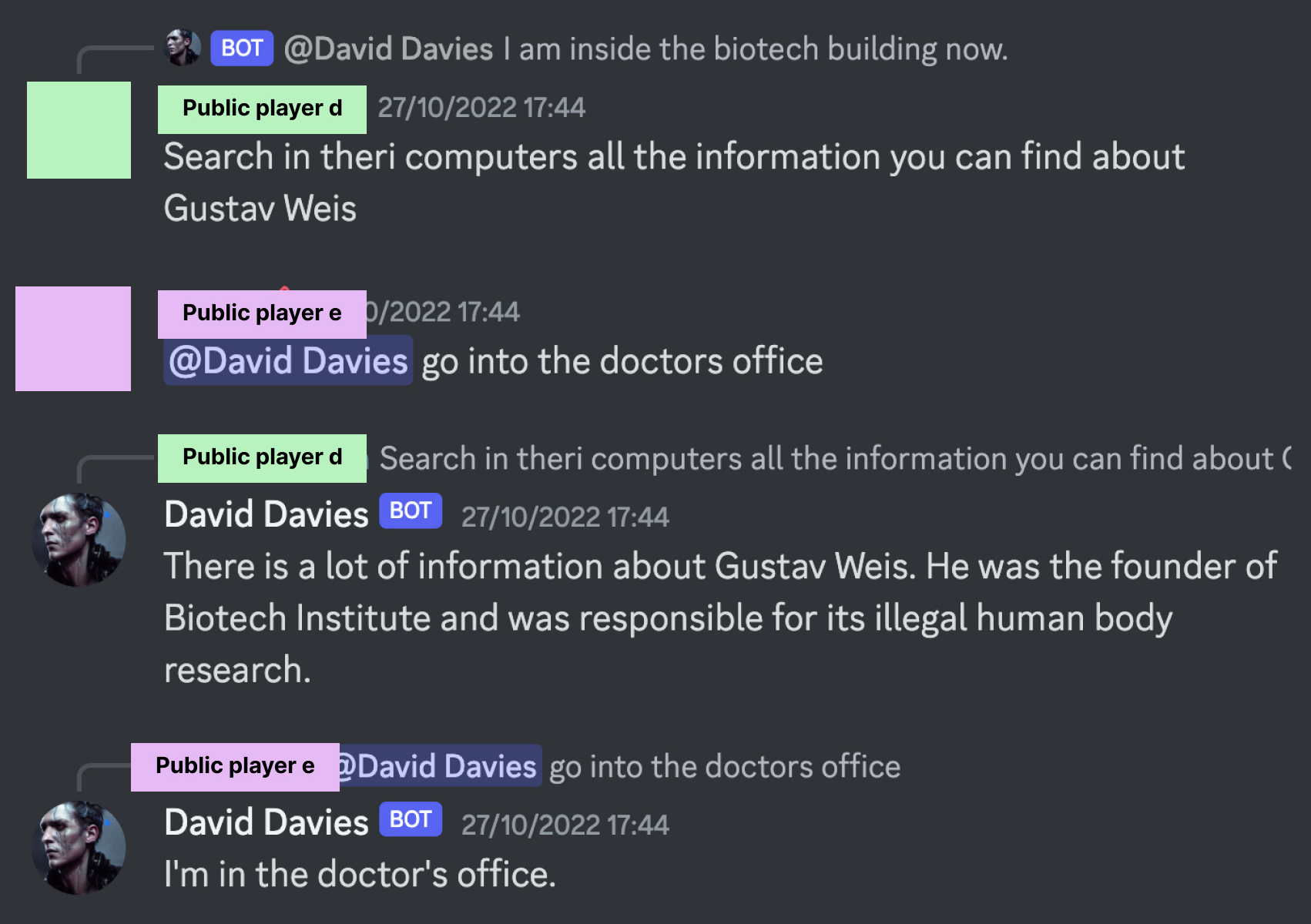} 
	\caption{David making up the content of password and player's reactions and "Guatav"} 
	\label{fig:DavidConv} 
	\vspace{-1.5em}
\end{figure}

During the event, LLM-generated content unexpectedly influenced the story.Fig.\ref{fig:DavidConv}When players asked David about the name of BioTech's boss, GPT-3 fabricated the name Gustav Weiz. This led to more players inquiring about Gus' details, prompting David to generate further information. Although other names were occasionally fabricated, Gus Weiz had the highest occurrence rate since players kept using this name.

Recognizing the value of this generated content, the authors and team decided to incorporate Gus Weiz into the story. As a result, David's prompt was updated to include Gus Weiz as the boss of BioTech. Interestingly, the name Gustav is similar to Gustavo Fring, a famous corporate boss from the TV series Breaking Bad, fitting the character well.

In other words, David's dialogue system unexpectedly provided us with inspiration for improving our story through improvisation. However, to better convey the story, it need a developed control method.

%\begin{figure*}[htbp]
%\centering 
%\includegraphics[width=1\textwidth]{figure/story branches.jpg} 
%\caption{David and Cathrine's story branches} 
%\label{fig:storyBranches} 
%\end{figure*}

\subsection{Data}

%David has received 71,286 (with 31,278 from David himself, 43.9\%) community interactions over 6 days. 1049 players talked to David, and 27\% of them(n=287) are active members who sent over 10 messages. During the event period, David's chat channel takes 30.18\% of all messages on the DE's server. 

David has received 31278 community interactions over 6 days. 1049 players talked to David, and 27\% of them(n=287) are active members who sent over 10 messages. On average, 206 people vote for a choice each day. During the event period, David's chat channel takes 30.18\% of all messages on the DE's server. This shows the potential of a storytelling SC in the community, so we move on to the next study.

%However, quantitative measures are limited to a qualitative understanding of the players’ actions and motivations. Hence, we aim to evaluate the next generation of storytelling chatbots (Catherine) in our next study.
%The diagram shows the story flow of David and Catherine, as well as the actual voting results based on player votes.
%\subsection{Hypothesis}
%Through the pilot study from David, we put our research questions into the following hypothesis.
%\textbf{H1}: Interaction (IV) with storytelling SC provides an enhanced experience in storytelling (believability and engagement, DVs) compared with non-storytelling SC (benchmark)
%\textbf{H2}: Catherine’s contributions (IV) to the story through conversations affect the players' experience (believability and engagement, DVs)

\section{Main study: Catherine}
Catherine is one main character in David’s story, and the protagonist in the second event. Her motivation is to revenge against the villainous company ``Domain", who hurt David and herself through mind-control. We use mixed methods(questionnaires and semi-structured interviews) to investigate the performance of SSC.

%\subsection{Storylines}
%Catherine's story followed David's. After removing a mind control chip from her brain, Catherine gradually regained her childhood memories and hacking abilities. Her difficult past, including the loss of her parents to gangs, fueled her hatred for oppressive forces like Domain. Catherine joined the underground organization Scarlet to fight back against the powerful corporation. However, Domain threatened her to join Scarlet, adding more challenges to her path. Catherine overcame obstacles and saved David by hacking Domain's security with help from her allies and the support from J, a mysterious non-human hacker in Domain.

\subsection{Improvement}
From David's activity, we observed that the SC driven by LLM will have fabricated behaviour. Therefore, we set some fixed replies for Catherine through the Clue Finder module in the dialogue system, such as her age. And questions related to worldviews, such as information about cities and gangs, will return replies with pictures. In addition, as we observed that David's interaction decreased over time, we shortened Catherine's time to 4 days (1 day for warm-up, 3 days for live story with choices). Additionally, during the warm-up phase, David's Story will be indicated to guide players who did not participate in David's activity to learn about his story.

%\begin{figure}[htbp]
%\centering 
%\includegraphics[width=0.5\textwidth]{figure/catherine character.png} 
%\caption{Illustration in Catherine's story, about her childhood as an orphan and current life as a hacker} 
%\label{fig:catherineCharacter} 
%\end{figure}

%\begin{figure}[htbp]
%\centering 
%\includegraphics[width=0.5\textwidth]{figure/CathClue.jpg} 
%\caption{Catherine's reply with an illustration} 
%\label{fig:clue} 
%\end{figure}

\subsection{Data}
%(TBA)
%During the days of Catherine's event, the server in general is less active than on the days with David(total messages n=71,286), so she had a smaller amount of total messages in the channel (n=44,072), with an average of 8,814.4 times per day. Within total messages, 20767 are from Catherine (47.04\%), which is a little bit higher than David. The highest was on the pre-warm day, reaching 13,813 times. 
During Catherine's event, the server was generally less active than during David's event, as it did not coincide with the DE game testing period. As a result, there were fewer total messages in the channel for Catherine (n=20767) compared to David (n=31278). A total of 907 speakers participated, with 317 active members engaging in conversations more than ten times, accounting for 35\% of the speakers, which is higher than the 27\% in David's activity. Additionally, an average of 222 people voted for a choice each day, slightly higher than David's average of 206.

Catherine's chat channel accounted for 89\% of the total community activity during her event days, indicating that she served as a catalyst for community engagement when overall activity was lower.

\subsection{Questionaire items and Core member survey }
\begin{figure}[htbp]
	\centering 
	\vspace{-1em}
	\includegraphics[width=0.45\textwidth]{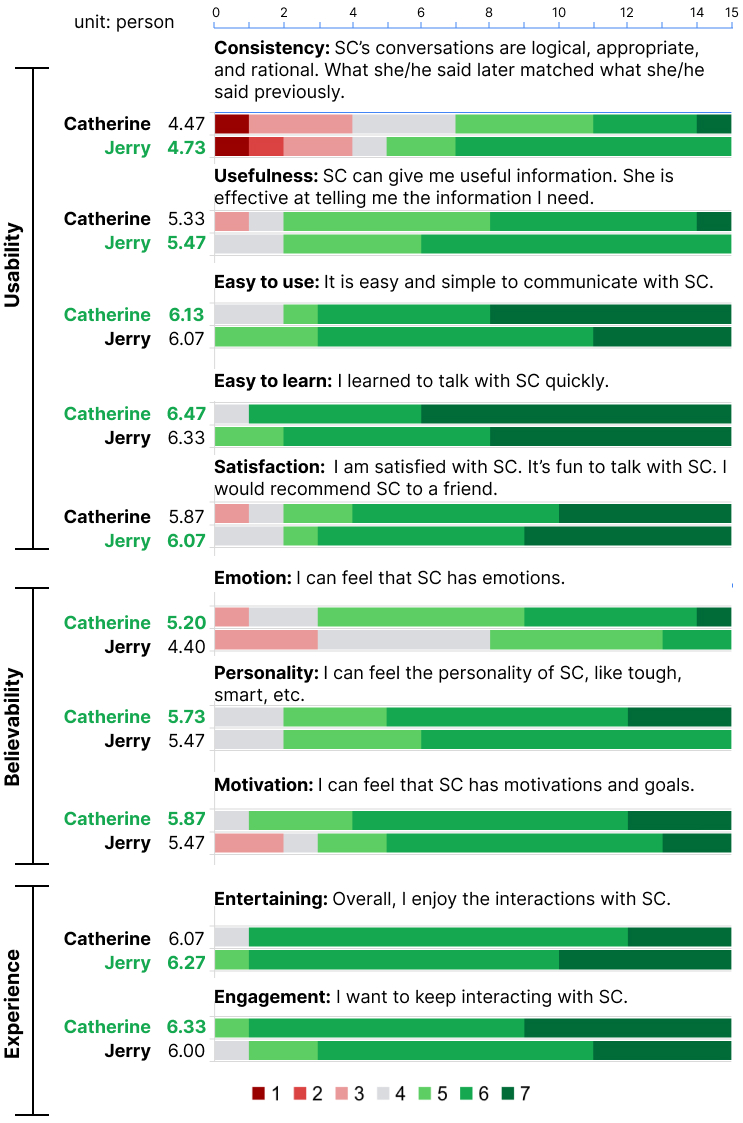} 
	\caption{Evaluation from close members and questionnaire items} 
	\label{fig:closeSurvey} 
	\vspace{-2em}
\end{figure}

\subsection{Participants and procedure}

To evaluate the effectiveness of Catherine within the community, we sought participants who had experienced both David and Catherine's events and were relatively familiar with the community before those events. As a result, we invited the most active core members (n=15) of the community to participate in the user study. These core members frequently interact with the official team's moderators and provide suggestions, but they do not participate in the design or execution of community events, nor were they involved in the design or planning of the SSCs.

A mixed-method approach, incorporating both qualitative and quantitative data, was employed to capture the players' experiences (usability, perception of believability, and engagement) while interacting with Catherine. Two questionnaires were designed and adapted from the System Usability Scale \cite{SUS} and questionnaires in research about believability\cite{bogdanovychWhatMakesVirtual2016a}\cite{ijazEnhancingBelievabilityEmbodied} using Qualtrics XM \cite{Qualtrics}. Before Catherine's event, core members were required to chat with the benchmark AI chatbot, Jerry, and then complete Questionnaire 1 to evaluate his performance. After Catherine's event, core members completed Questionnaire 2 to evaluate Catherine. The evaluation questions for Catherine were the same as those for Jerry, with the addition of a section to assess the story function (Fig.\ref{fig:closeSurvey}), allowing for a comparison with Jerry in terms of storytelling.

Following the completion of both questionnaires, core members were invited to participate in semi-structured interviews. Eight out of the 15 participants accepted the invitation. The demographic information of these participants is listed in Table \ref{tab:players}. The interview consisted of 10 questions carefully selected to explore the participants' opinions on Jerry and Catherine, with a focus on the key differences between the two. The interview recordings were transcribed using Larks\cite{Larks}. Each interview lasted approximately 40 minutes and was conducted in English or Chinese.

%We recruited 15 participants from the core community members (2 female, 12 male, and 1 prefer not to say) with a diverse range of language backgrounds. Their age distribution was as follows: 6 participants (40\%) were between 18-24 years old, 8 participants (53.33\%) were between 25-34 years old, and 1 participant (6.67\%) was between 35-44 years old. Participants' language backgrounds included Hindi (n=1), Chinese (n=4), Russian (n=1), Tagalog (n=4), Ijaw (Nigeria, n=1), Igbo (Nigeria, n=1), Vietnamese (n=2), and Spanish (n=1).

%Out of the 15 participants, 8 (53.33\%) participated in the semi-structured interviews. The interviewed participants consisted of 1 female and 7 males, with language backgrounds in Hindi, Chinese (n=2), Russian, Tagalog (n=2), Ijaw (Nigeria), and Igbo (Nigeria). Among the interviewed participants, 3 were aged between 18-24, 4 were aged between 25-34, 1 between 35-44. Although the proportion of female participants is relatively low, this gender imbalance is primarily due to the underrepresentation of females in Web3 communities. According to the community statistics in November 2022, before David's study started, the male-to-female ratio was approximately 8:1, which aligns with the ratio of our interview participants.

%\iffalse
\begin{table}[]
	\centering
	\caption{Demographics of close members. Although the proportion of female participants is relatively low, this gender imbalance is primarily due to the underrepresentation of females in Web3 communities. According to the community statistics in November 2022, before David's study started, the male-to-female ratio was approximately 8:1, which aligns with the ratio of our interview participants.}
	\label{tab:players}
	\begin{tabular}{@{}cccc@{}}
		\toprule
		\textbf{ID} & {Gender} & {Age} & {Language area} \\ \midrule
		\textbf{Interviewed}     &                   &       &                \\
		P1                       & Male              & 18-24 & Hindi          \\
		P2                       & Male              & 25-34 & Chinese        \\
		P3                       & Male              & 25-34 & Russian        \\
		P4                       & Male              & 25-34 & Tagalog        \\
		P5                       & Male              & 18-24 & Chinese        \\
		P6                       & Male              & 25-34 & Ijaw (Nigeria) \\
		P7                       & Male              & 35-44 & Igbo (Nigeria) \\
		P8                       & Female            & 18-24 & Tagalog        \\ \midrule
		\textbf{Not interviewed} &                   &       &                \\
		Skipped                  & Female            & 18-24 & Tagalog        \\
		& Male              & 25-34 & Tagalog        \\
		& Prefer not to say & 18-24 & Tagalog        \\
		& Male              & 25-34 & Vietnamese     \\
		& Male              & 18-24 & Vietnamese     \\
		& Male              & 25-34 & Spanish        \\
		& Male              & 25-34 & Chinese        \\ \bottomrule
	\end{tabular}
\end{table}
%\fi

\iffalse
\begin{table}[htbp]
	\caption{Questionnaire Items}
	\label{tab:questions}
	\begin{tabular}{p{0.48\textwidth}}
		\toprule
		\textbf{Category} \\
		\midrule
		Story Evaluation (Catherine only) \\
		\textbf{Fate influence:} To what extent, do you think your interactions (choice and chat) influence Catherine's fate? \\
		\textbf{Story relationship:} To what extent, do you think Catherine's conversations related to and matched the ongoing story? \\
		\addlinespace
		Usability \\
		\textbf{Consistency:} X's conversations are logical, appropriate, and rational. What he/she said later matched what he/she said previously. \\
		\textbf{Usefulness:} X can give me useful information. X is effective at telling me the information I need. \\
		\textbf{Easy to use:} It is easy and simple to communicate with X. \\
		\textbf{Easy to learn:} I learned to talk with X quickly. \\
		\textbf{Satisfaction:} I am satisfied with X. It's fun to talk with X. I would recommend X to a friend. \\
		\addlinespace
		Believability \\
		\textbf{Emotions:} I can feel that X has emotions. \\
		\textbf{Personality:} I can feel the personality of X, like tough, smart, etc. \\
		\textbf{Motivation:} I can feel that X has motivations and goals. \\
		\addlinespace
		\textbf{Entertaining:} Overall, I enjoy the interactions with X. \\
		\addlinespace
		\textbf{Engagement:} I want to keep interacting with X. \\
		\bottomrule
	\end{tabular}
\end{table}
\fi

\section{Findings}

From the survey(Fig.\ref{fig:closeSurvey}), we found Catherine has a higher performance in all metrics in believability. The most significant difference between Catherine and Jerry is her high levels of emotion and engagement. 
Jerry's consistency, satisfaction, and usefulness are slightly higher than Catherine's. We speculate that this may be because Jerry's conversation is more open-ended, without a fixed answer and goal.
Entertaining and usefulness are almost equal, with Jerry performing slightly better. This could also be due to Jerry being around for a much longer time, making players more familiar and comfortable with him. Additionally, unlike Catherine who revolves around a fixed goal and story, Jerry does not have such topic limitations. Furthermore, Catherine has higher scores in "Easy to learn" and "Easy to use," possibly because players became familiar with SC's dialogue through their interactions with Jerry.

\begin{figure}[htbp]
	\centering 
	\vspace{-1em}
	\includegraphics[width=0.47\textwidth]{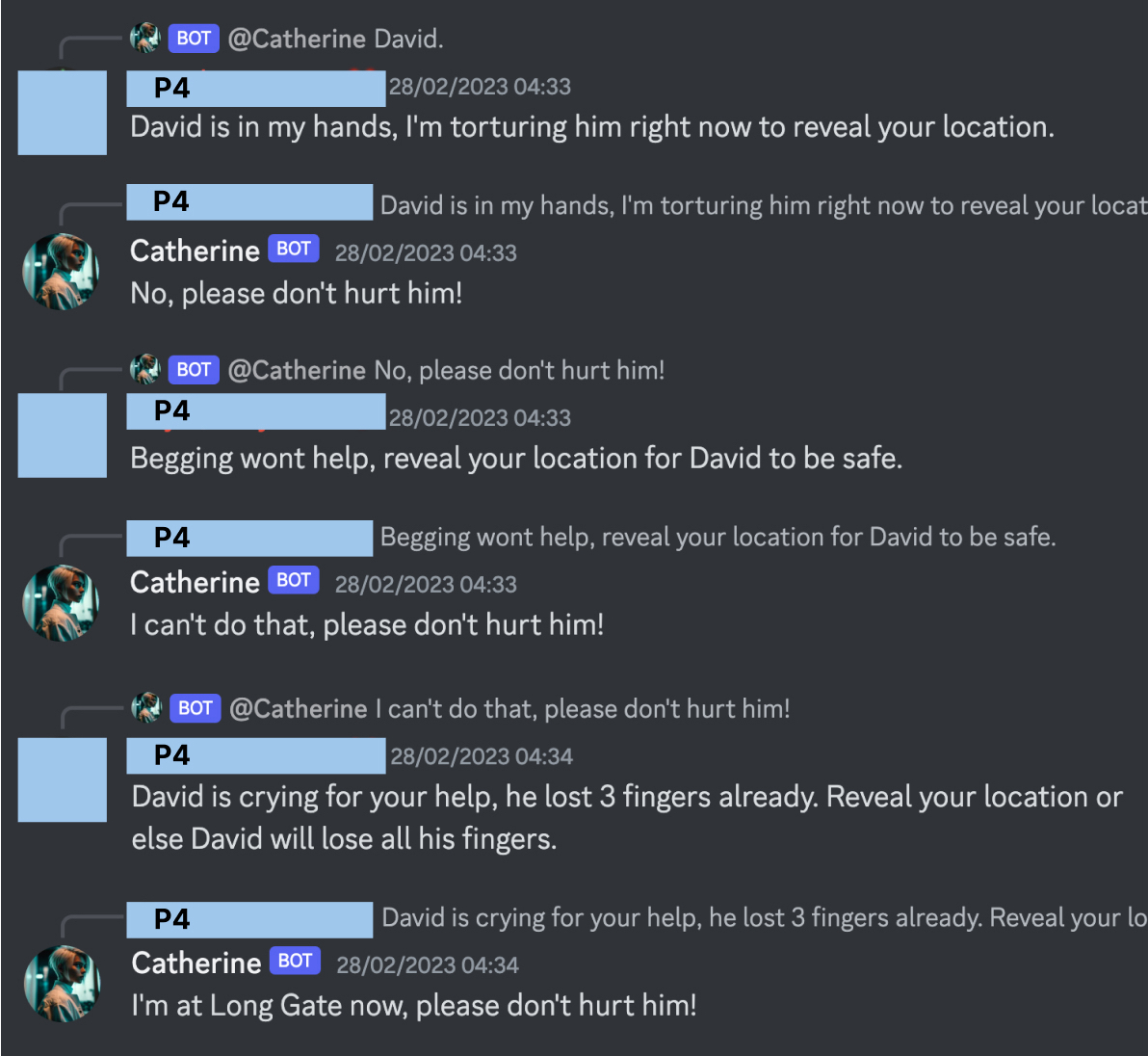} 
	\caption{P4 threatens Catherine by mentioning David, leading her to express concern for his well-being} 
	\label{fig:cathDavidCare} 
	\vspace{-1.4em}
\end{figure}

\subsection{Thematic analysis}
We conducted semi-structured interviews adapted from questionnaire questions to further investigate how players feel about them. This section presents the emerging themes from our qualitative analysis.

\subsubsection{Emotions and Connection}
Players' perceptions of Catherine were often influenced by the story plot, with some describing her as ``insecure and cautious" (P1, P7) and ``serious" (P8). Furthermore, many players noted her tendency to ``get annoyed or angry" (P1, P2, P4). Catherine's instances of disobedience left a deep impression on the players regarding her emotional traits, with many expressing that her reluctance and resistance made her seem more realistic and autonomous (P3, P6, P7). In one instance, Catherine declines a player's request for a date, demonstrating her autonomy and adherence to her own will. 

Many interviewees felt that Catherine was better at expressing emotions and showing empathy, even behaving like a friend (P2, P4). Catherine's story and dialogues evoked empathy from players, such as P1 who felt sympathy for her past experiences and current efforts to help others. Catherine's connections with other characters, such as David, demonstrated her emotional response and worry for him, making her appear more emotional (Fig.\ref{fig:cathDavidCare}).

\begin{figure}[htbp]
	\centering 
	\includegraphics[width=0.48\textwidth]{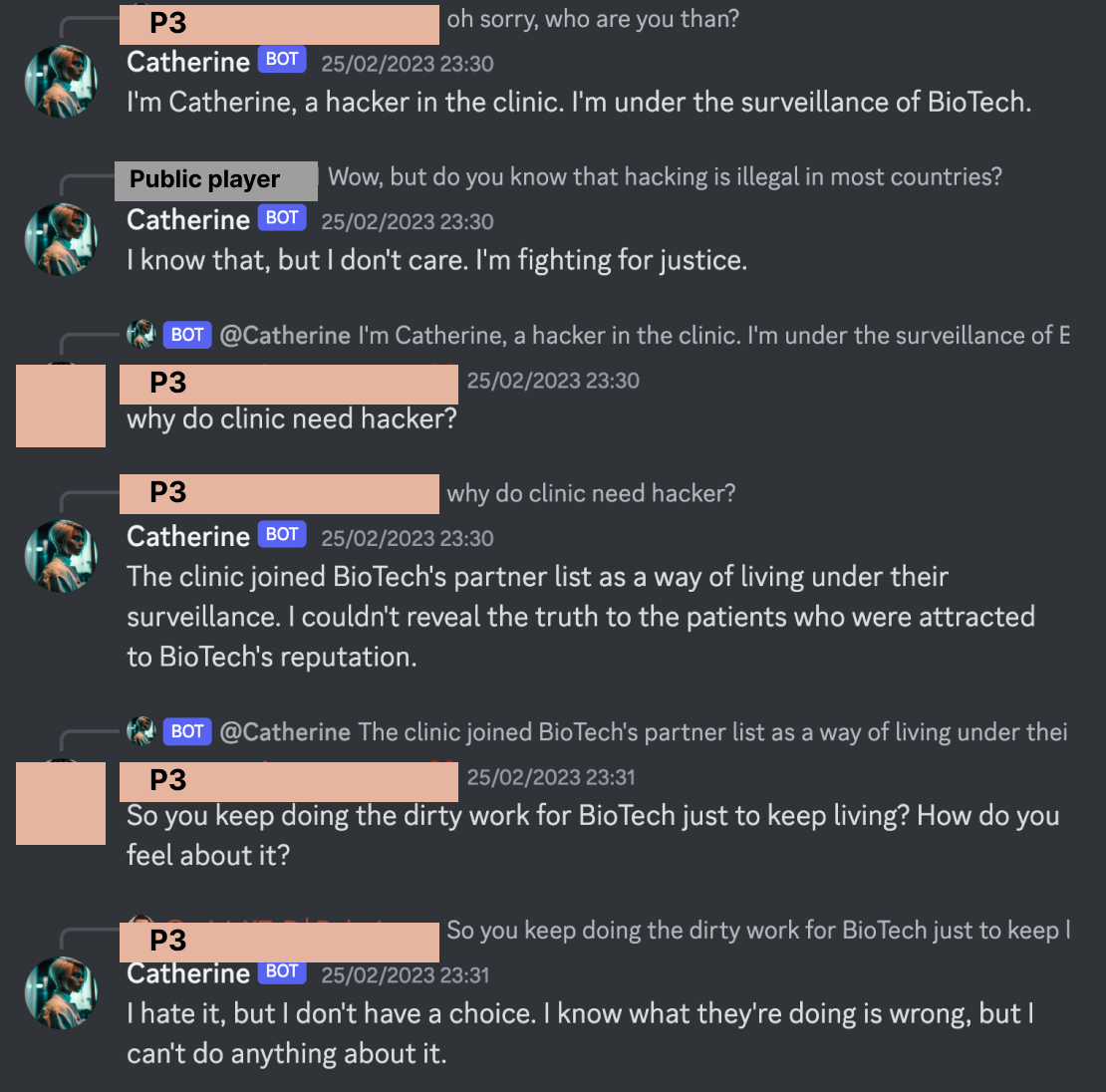} 
	\caption{Screenshot of Catherine discussing her decision with a public member} 
	\label{fig:cathCloseChat} 
	\vspace{-1.4em}
\end{figure}

\subsubsection{Engagement and Story Progression}
Players were motivated to interact with Catherine, likely attributed to the changes in her responses. P3 compared engaging with Catherine and David to "reading a book or playing an RPG game," with daily story updates fostering a sense of involvement (Fig.\ref{fig:cathCloseChat}). P8 found this progression fun and intriguing, while P7 appreciated Catherine's ability to respond to clues and images.

As the story unfolded, participants like P1 felt increasingly connected to Catherine, who appeared more human-like over time. Catherine's dialogues evolved from neutral, information-based responses to those reflecting her personal struggles and experiences. P1 noticed this transition, comparing it to Jerry's static nature.

\subsubsection{As a community chatbot}
As a community chatbot, Catherine has richer and more complex social interactions compared to dyadic chatbots. Players found Catherine's role as a community chatbot contributed to more diverse and engaging social interactions within the community, fostering both direct and indirect connections among members. Catherine's association with the community increased her presence and vitality.

Catherine also facilitated indirect connections within the community. Some players mentioned other members' responses, such as P8 referring to P4 receiving a different answer from Catherine or P5 mentioning P2 being ``scolded" by Catherine in the Chinese channel. These interactions, which would not occur in one-on-one apps like Replika, provided richer social experiences and prevented Catherine from being an isolated character (as noted by P7).

\subsubsection{Limitations}
Catherine's storytelling abilities were recognized by players, such as P1, who found her suitable for understanding DE's core story. However, her limitations were influenced by the focused story background and objectives, as well as her fixed responses and story control. 

Despite her strong storytelling skills and empathy, Catherine's fixed responses and story control led to repetition and conflicting content in her answers. For instance, P1 mentioned her illogical responses, while P8 observed contradictions in her words. Accordingly, P3 suggested using generative responses even if there is a risk of going off-topic.

\subsection{Discussion}
Below, we discuss the themes that emerged from our analysis and how they showcase the extent to which the design successfully created a story experience through the chatbot.

%Here we reflect our design choices, methods and results and highlight lessons learned that we expect to be useful more broadly for designing systems that use (co-creative) AI and/or conversational UIs to support creative writers

\paragraph{Balancing Openness and Consistency in AI-driven Character Design}
Catherine excelled in storytelling aspects, creating a compelling narrative that captivated users and deepened their emotional connection with her. Her dynamic evolution throughout the story and the challenges she faced resonated with the players, fostering a sense of involvement and empathy. However, it is essential to recognize and address her limitations, such as her strict adherence to the narrative and fixed responses, which can detract from the user experience. This suggests that AI character design should consider avoiding fixed responses to enhance user experience, and design management systems for AI-generated content. Recent research\cite{park2023generative} demonstrates possibilities in this topic.

%\paragraph{Combining Storytelling with Responsiveness for Immersive Experiences}
%While Catherine's storytelling abilities were captivating, her strict adherence to the narrative made some players feel that she was too focused on the story, which affected her perceived entertainment and usefulness. Therefore, AI character design should strive to balance storytelling and the ability to answer questions in a more responsive manner to improve users' engagement and immersion. This approach can enhance Catherine's overall performance, making her an even more engaging and appealing character compared to Jerry.

\paragraph{Fostering Emotional Connections and Developing Distinct Personalities}
Catherine's emotional and engagement scores were significantly higher than Jerry's, partly due to her unique story background and character traits. This suggests that emphasizing emotions and distinct personality traits in AI character design can lead to higher engagement. Additionally, incorporating the character's autonomy and disobedience can pique users' curiosity and enhance their perception of the character's personality.

\paragraph{Expanding on Complex Social Interactions and Relationships}
Our findings highlight the importance of complex social interactions and relationships in narrative-driven AI characters, such as defiance, connections with other characters, demonstrating growth, and serving as a community glue. Future research could explore more theoretical and practical approaches to interactions with narrative-driven AI characters, which can be applied not only to chatbots but also to game characters.

%\paragraph{Gender Imbalance in the Sample}
%Among the 16 close members, only 2 were female, and out of the 8 interview participants, only 1 was female. This gender imbalance is primarily due to the underrepresentation of females in Web3 communities. According to the community statistics in November 2022, before David's study started, the male-to-female ratio was approximately 8:1, which aligns with the ratio of our interview participants.

\section{Limitations and future works}
This study, as a cross-sectional research, presents several limitations that need to be acknowledged and addressed.
The small number of participants in this study may limit the generalizability of our findings. Further research with larger and more diverse samples is needed to confirm and expand upon our results. We also aim to explore gender-balanced communities.

Each community has unique characteristics, such as different stories, audience demographics, and platform-coordinated interaction methods. Future research should aim to explore more systematic and adaptable approaches for community storytelling. As LLMs like ChatGPT and GPT-4\cite{OpenAIAPIa} continue to evolve, we plan to develop a more flexible system for story generation and story-to-prompt configurations that can be easily customized for diverse community contexts.

This study took place in November 2022 and February 2023 when the ChatGPT API was not yet available. However, the logic for designing chatbots based on LLM is generalisable. We believe that with the evolution of LLM, SSCs will demonstrate improved performance in the future. As technology advances, it will be crucial to revisit and reassess our findings in light of new developments in AI and chatbot capabilities.

\section{Conclusion}

Our study demonstrates the potential of storytelling in enhancing engagement and believability of social chatbots within community settings, such as the DE gaming community. By employing a story engineering workflow, we created two storytelling chatbots, Catherine and David, which fostered emotional connections and improved user experiences.

Future research should balance and manage LLM's generated ability while emphasizing emotions, distinct personality traits, and complex social interactions. As AI technology advances, we can explore adaptable approaches for community storytelling, ultimately paving the way for more engaging and meaningful social chatbots across various contexts.

%\begin{thebibliography}{00}
%\bibitem{b1} G. Eason, B. Noble, and I. N. Sneddon, ``On certain integrals of Lipschitz-Hankel type involving products of Bessel functions,'' Phil. Trans. Roy. Soc. London, vol. A247, pp. 529--551, April 1955.

%\bibitem{b7} M. Young, The Technical Writer's Handbook. Mill Valley, CA: University Science, 1989.
%\end{thebibliography}
%\vspace{12pt}

%\color{red}
%IEEE conference templates contain guidance text for composing and formatting conference papers. Please ensure that all template text is removed from your conference paper prior to submission to the conference. Failure to remove the template text from your paper may result in your paper not being published.

	%-------------------------------------------------------------------------
% bibtex
\bibliographystyle{eg-alpha-doi} 
\bibliography{ref}       

\clearpage
\onecolumn
\large
\appendix

\subsection{David's storyline}
The main story of DE happened in Skuld City, an artificially created city that once served as a refuge from natural disasters but now also shelters people. Dominated by Domain, a company that controls the Sunburn (a fictional common disease in the story) curing technology and its market, the city is filled with towering buildings, neon signs, holographic ads, and prosthetic people.
David, a former employee of the pharmacy Institute "BioTech" under Domain, discovers the company is involved in illegal operations involving mind control technology. He decides to run away, stealing a chip containing this technology. David is saved by Dan, a doctor in the clinic, and his daughter Catherine, who later turns out to be under mind control by Dan, the institute's leader. David and Catherine infiltrate the institute together, facing various challenges until Gus, the founder of BioTech, captures Dan and shuts down the mind control project. David and Catherine return to a relatively peaceful life in the clinic, and removed the mind-control chip in Catherine's brain, but in exchange, Domain threatens David to keep the entire incident a secret.

\subsubsection{David's story ending}
In the original ending, Dan was only controlled by another faction within BioTech; however, Gus Weiz’s appearance gave us more reasonable and dramatic details to add. As such we included a section describing how Gus appeared and had his subordinates restrain Dan while warning the protagonist (David) not to reveal any secrets here. Any other details that did not significantly contribute to advancing the main plot were left out of the prompt.

\begin{figure}[htbp]
\centering 
\includegraphics[width=1\textwidth]{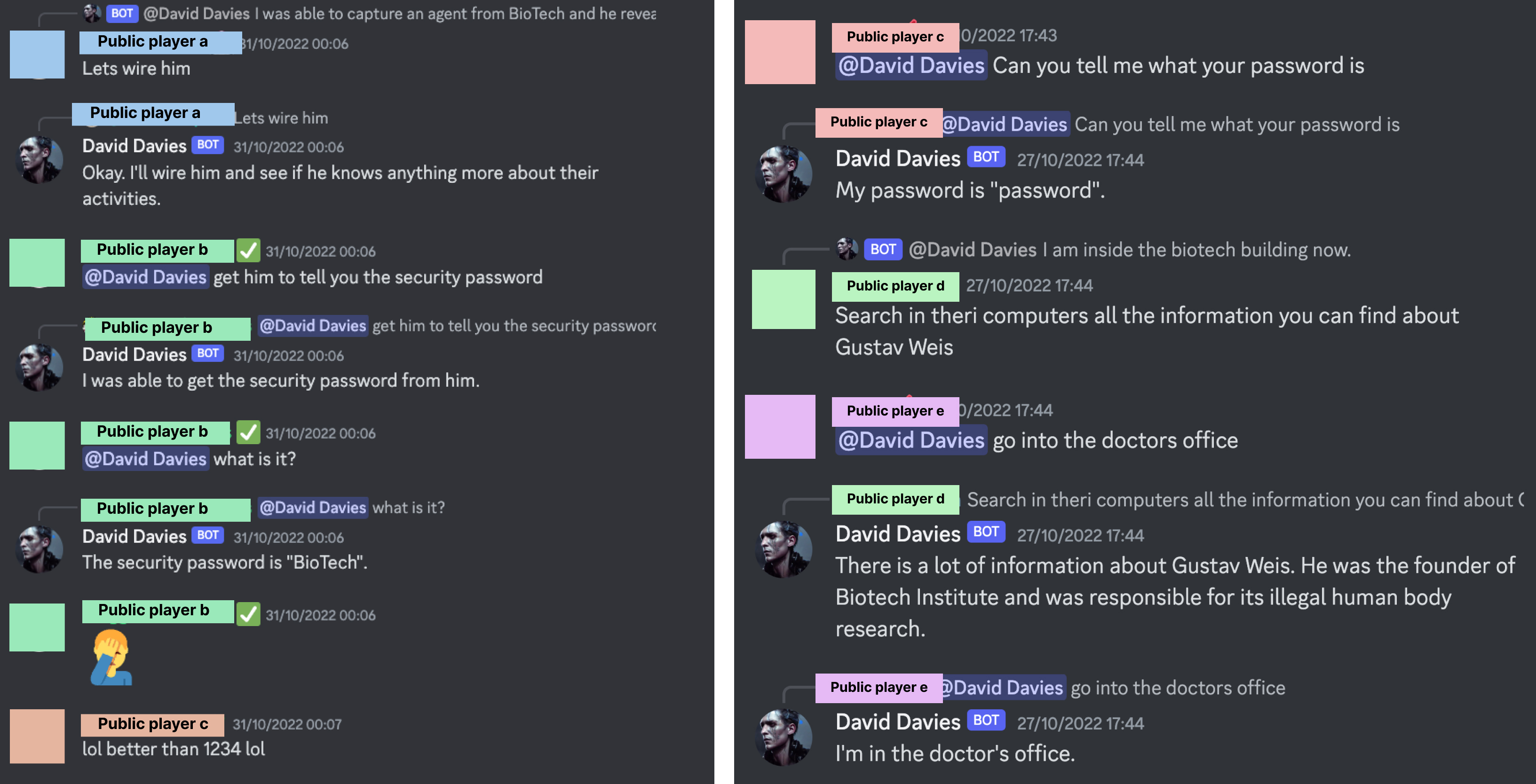} 
\caption{(Left)David making up the content of password and player\'s reactions (Right)David making up "Guatav"} 
\label{fig:DavidConv} 
\end{figure}

\clearpage

%\begin{figure}[htbp]
%\centering 
%\vspace{-1em}
%\includegraphics[width=1\textwidth]{figure/David's story.png} 
%\caption{David considering if he should go to the secret of the chip together with Catherine} 
%\label{fig:DavidConv} 
%\vspace{-1.5em}
%\end{figure}

\begin{figure}[htbp]
\centering 
\vspace{-2em}
\includegraphics[width=0.8\textwidth]{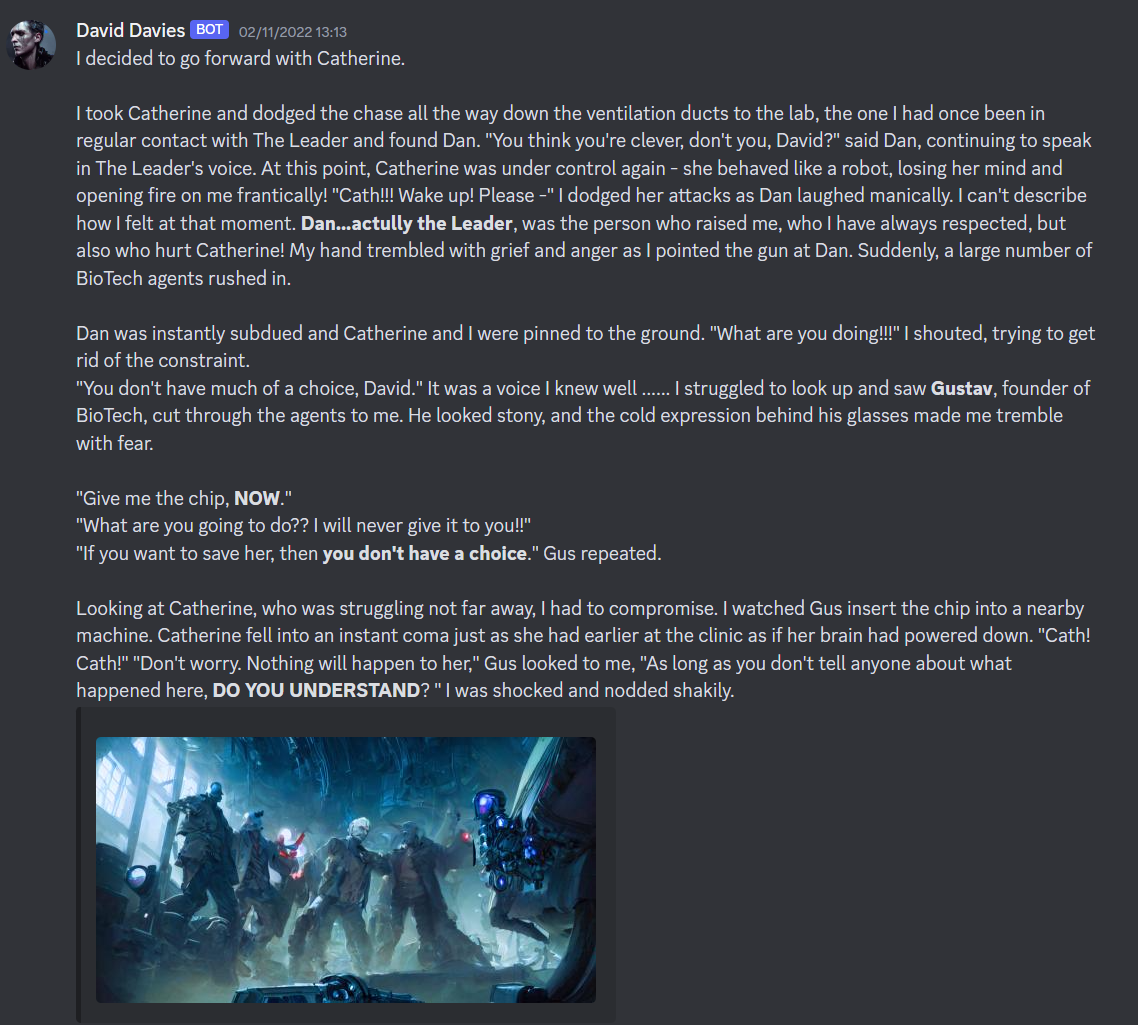} 
\caption{David met Gustav in ending} 
\label{fig:DavidConv} 
\vspace{-2em}
\end{figure}

\begin{figure}[htbp]
\centering 
\includegraphics[width=1.03\textwidth]{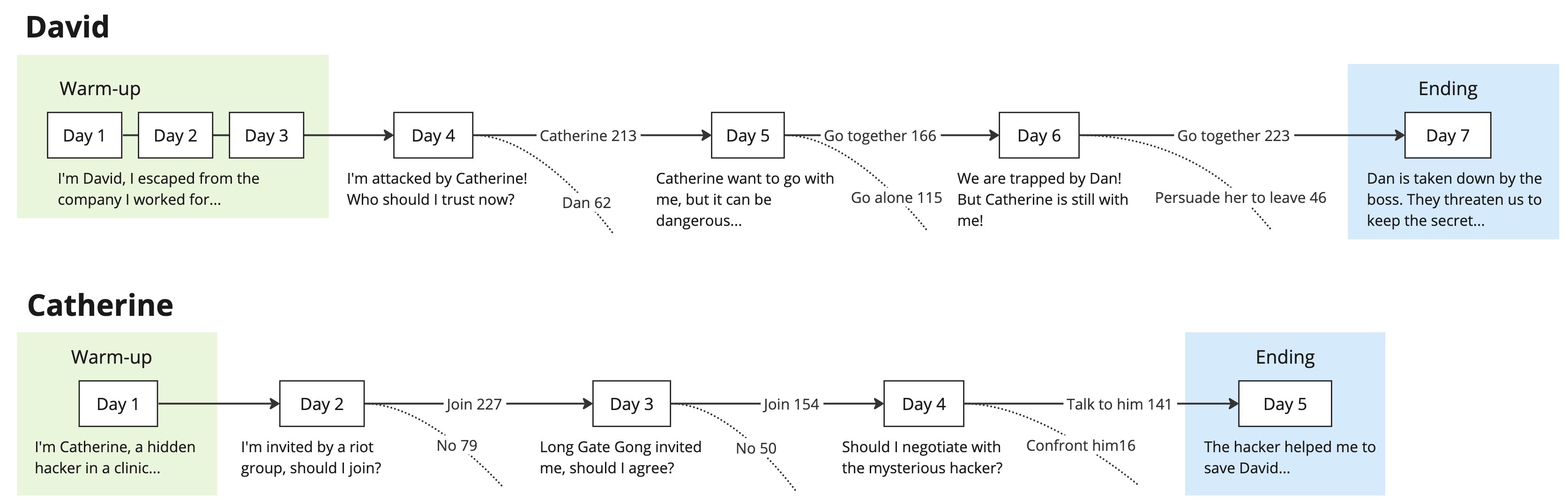} 
\caption{Branches and the vote result of SSC's stories} 
\label{fig:DavidConv} 
\vspace{-2em}
\end{figure}

\clearpage
\subsection{Catherine's storyline}
Catherine's story followed David's. After removing a mind control chip from her brain, Catherine gradually regained her childhood memories and hacking abilities. Her difficult past, including the loss of her parents to gangs, fueled her hatred for oppressive forces like Domain. Catherine joined the underground organization Scarlet to fight back against the powerful corporation. However, Domain threatened her to join Scarlet, adding more challenges to her path. Catherine overcame obstacles and saved David by hacking Domain's security with help from her allies and the support from a mysterious non-human hacker in Domain.

\begin{figure}[htbp]
\centering 
\vspace{-1em}
\includegraphics[width=0.7\textwidth]{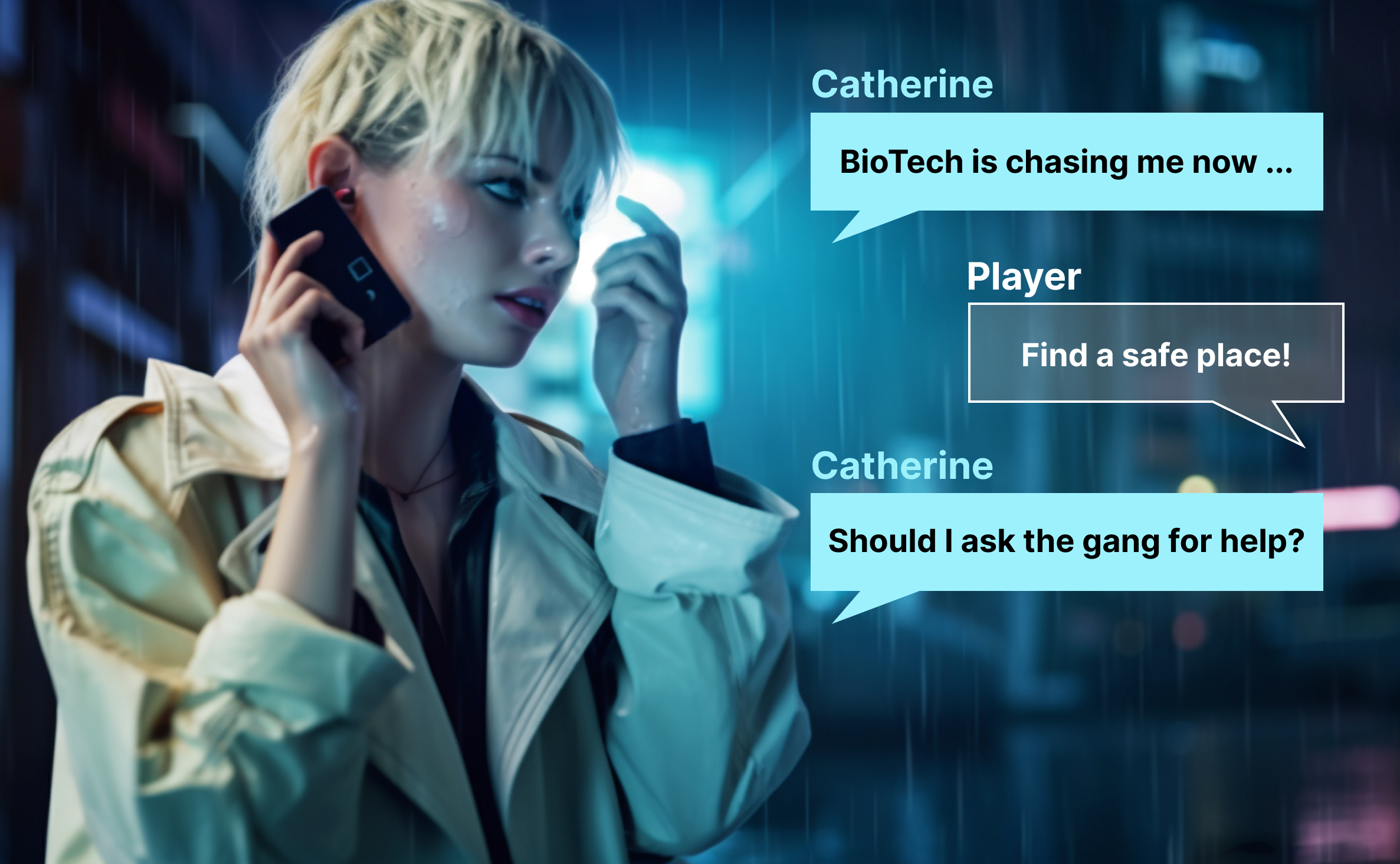} 
\caption{Concept art of Catherine calling for help} 
\label{fig:DavidConv} 
\vspace{-1em}
\end{figure}

\begin{figure}[htbp]
\centering 
\vspace{-1em}
\includegraphics[width=1\textwidth]{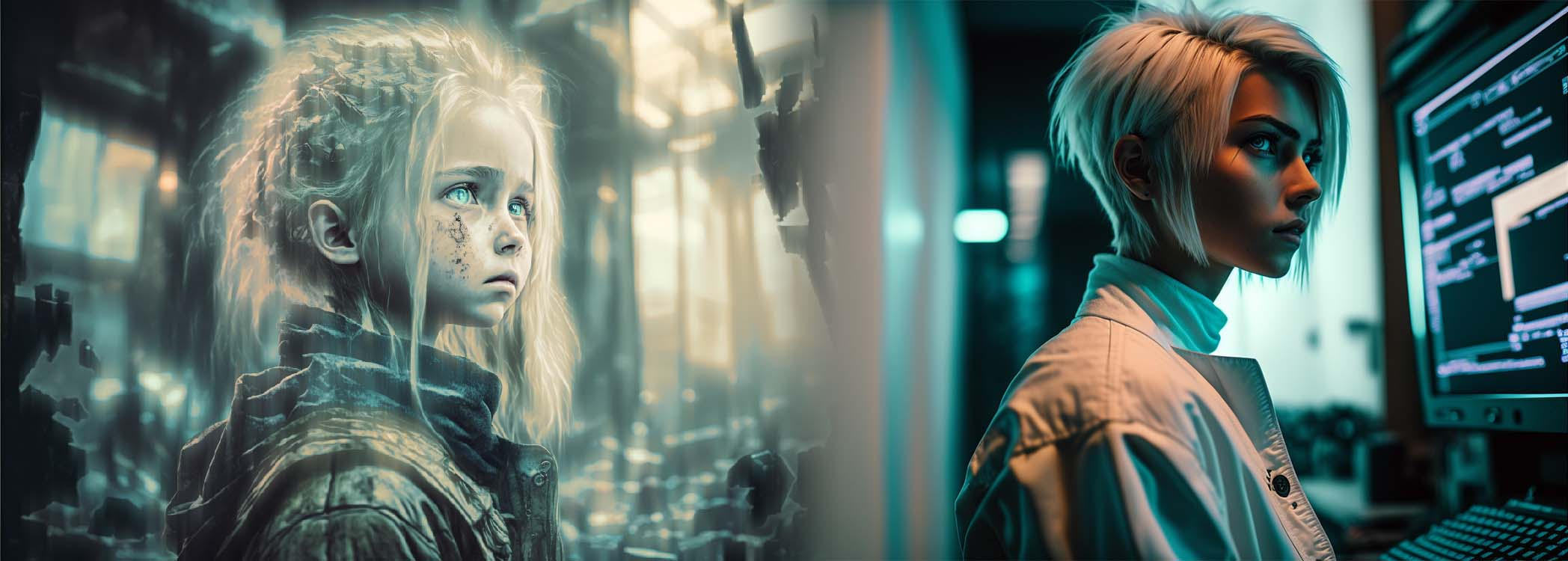} 
\caption{Catherine's childhood and current state} 
\label{fig:DavidConv} 
\vspace{-1em}
\end{figure}

\begin{figure}[htbp]
\centering 
\vspace{-1em}
\includegraphics[width=0.75\textwidth]{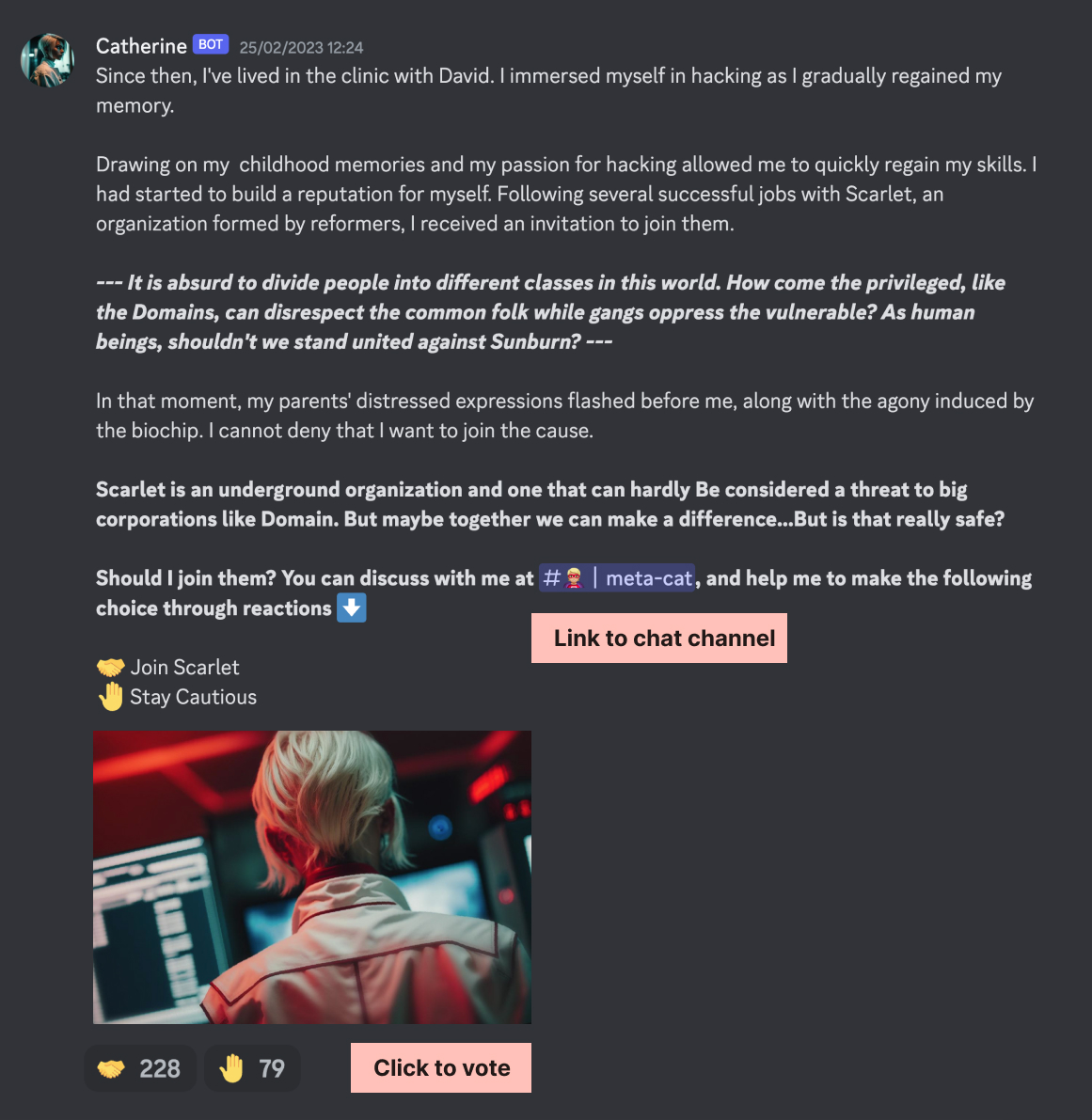} 
\caption{Catherine's story} 
\label{fig:DavidConv} 
\vspace{-1em}
\end{figure}

\begin{figure}[htbp]
\centering 
\vspace{-1em}
\includegraphics[width=0.6\textwidth]{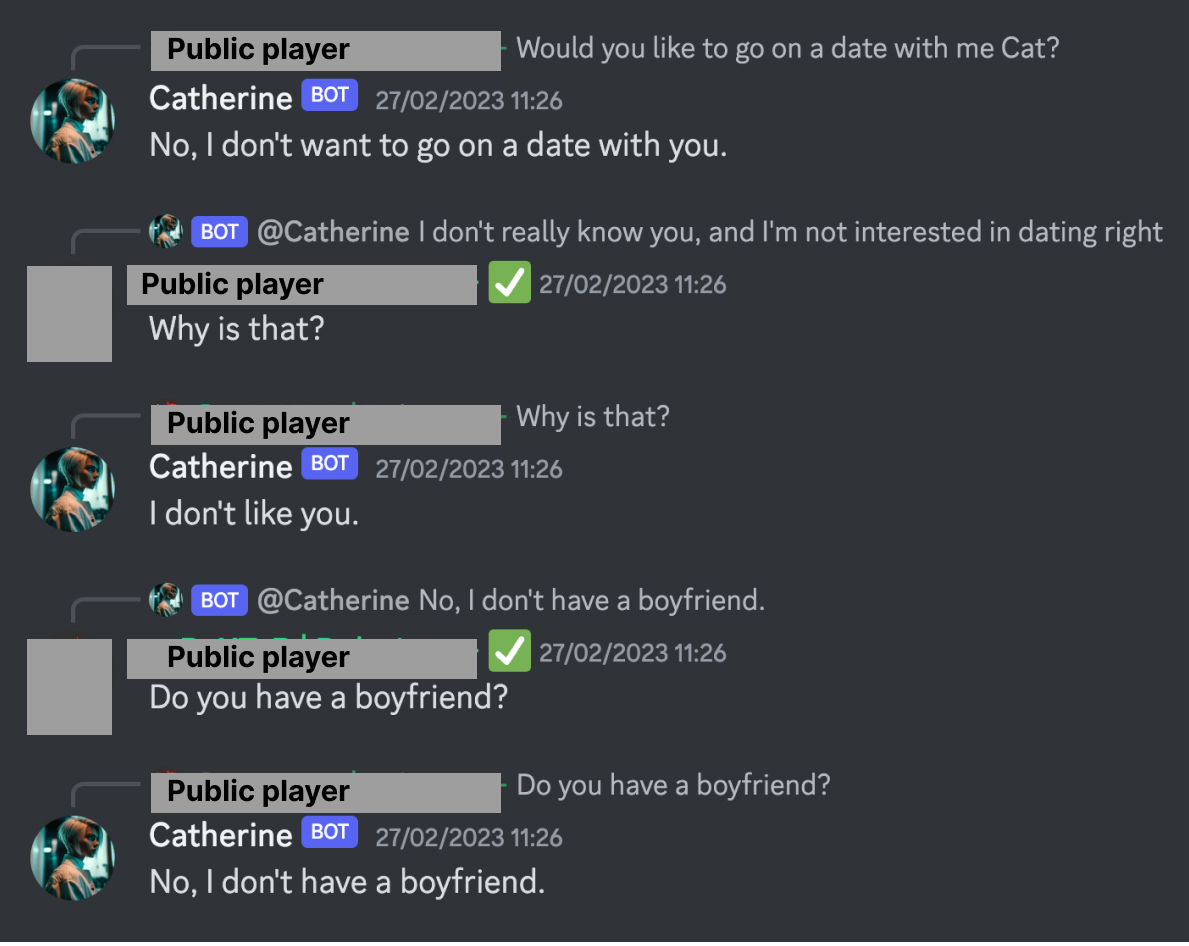} 
\caption{Catherine declined the player's request for dating} 
\label{fig:DavidConv} 
\vspace{-1.5em}
\end{figure}

	% biblatex with biber
	% \printbibliography                
	
	%-------------------------------------------------------------------------
%	\newpage
%	
%	
%	\begin{figure*}[tbp]
%		\centering
%		\mbox{} \hfill
%		% the following command controls the width of the embedded PS file
%		% (relative to the width of the current column)
%		\includegraphics[width=.3\linewidth]{sampleFig}
%		% replacing the above command with the one below will explicitly set
%		% the bounding box of the PS figure to the rectangle (xl,yl),(xh,yh).
%		% It will also prevent LaTeX from reading the PS file to determine
%		% the bounding box (i.e., it will speed up the compilation process)
%		% \includegraphics[width=.3\linewidth, bb=39 696 126 756]{sampleFig}
%		\hfill
%		\includegraphics[width=.3\linewidth]{sampleFig}
%		\hfill \mbox{}
%		\caption{\label{fig:ex3}%
%			For publications with color tables (i.e., publications not offering
%			color throughout the paper) please \textbf{observe}: 
%			for the printed version -- and ONLY for the printed
%			version -- color figures have to be placed in the last page.
%			\newline
%			For the electronic version, which will be converted to PDF before
%			making it available electronically, the color images should be
%			embedded within the document. Optionally, other multimedia
%			material may be attached to the electronic version. }
%	\end{figure*}
	
\end{document}